\pdfoutput=1

\documentclass[11pt]{article}

\usepackage[final]{acl}
\usepackage{times}
\usepackage{latexsym}

\usepackage[T1]{fontenc}

\usepackage[utf8]{inputenc}

\usepackage{microtype}

\usepackage{inconsolata}

\usepackage{graphicx}

\usepackage{booktabs} 
\usepackage{rotating}

\graphicspath{{img/}}

\title{Low-Resource Transliteration for Roman-Urdu and Urdu Using Transformer-Based Models}

\author{
  Umer Butt \\
  University of Saarland \\
  DFKI  \\
  Saarbrücken, Germany \\
  \texttt{mubu01@dfki.de} \\\And
  Stalin Varanasi \\
  University of Saarland \\
  DFKI  \\
  Saarbrücken, Germany \\
  \texttt{stalin.varanasi@dfki.de} \\\And
  Günter Neumann \\
  University of Saarland \\
  DFKI  \\
  Saarbrücken, Germany \\
  \texttt{guenter.neumann@dfki.de}
}

\begin{document}

\maketitle
\begin{abstract}
As the Information Retrieval (IR) field increasingly recognizes the importance of inclusivity, addressing the needs of low-resource languages remains a significant challenge. Transliteration between Urdu and its Romanized form, Roman Urdu, remains underexplored despite the widespread use of both scripts in South Asia. Prior work using RNNs on the Roman-Urdu-Parl dataset showed promising results but suffered from poor domain adaptability and limited evaluation.

We propose a transformer-based approach using the m2m100 multilingual translation model, enhanced with masked language modeling (MLM) pretraining and fine-tuning on both Roman-Urdu-Parl and the domain-diverse Dakshina dataset. To address previous evaluation flaws, we introduce rigorous dataset splits and assess performance using BLEU, character-level BLEU, and CHRF.

Our model achieves strong transliteration performance, with Char-BLEU scores of 96.37 for Urdu→Roman-Urdu and 97.44 for Roman-Urdu→Urdu. These results outperform both RNN baselines and GPT-4o Mini and demonstrate the effectiveness of multilingual transfer learning for low-resource transliteration tasks.

\end{abstract}

\section{Introduction}
Advancements in Natural Language Processing (NLP), particularly transformer-based architectures, have significantly improved various linguistic tasks, including machine translation, transliteration, and cross-lingual transfer learning. However, low-resource languages continue to face challenges due to data scarcity, limited model training resources, and domain adaptation issues. One such underrepresented language is Urdu, spoken widely in South Asia, which features a Perso-Arabic script, right-to-left writing direction, and rich morphology \cite{urdustats}.

In parallel to its standard script, a Romanized variant, known as Roman-Urdu, has become increasingly popular in digital communication, social media, and messaging platforms. This form of writing represents Urdu using the Latin alphabet, but lacks standardized spelling conventions, leading to high variability in the spelling of Romanized words. As a result, transliterating between Roman-Urdu and standard Urdu presents unique challenges, particularly in NLP applications. Despite its widespread use online, research on Roman-Urdu to Urdu or the opposite's transliteration remains scarce, limiting the development of language technologies that can effectively bridge the gap between these two writing systems.

Existing transliteration models for Roman-Urdu and Urdu primarily rely on Recurrent Neural Networks (RNNs) and Long Short-Term Memory (LSTM) architectures, as demonstrated in prior work using the Roman-Urdu-Parl (RUP) dataset \cite{Roman-urdu-parl}. While these models achieve reasonable performance, they exhibit limitations in handling complex linguistic structures, proper names, and cross-domain generalization. Furthermore, evaluation inconsistencies in prior research suggest that previously reported results may not fully reflect actual model performance. Additionally, models trained solely on Roman-Urdu-Parl struggle significantly when tested on other datasets such as Google's Dakshina dataset \cite{dakshina}, revealing domain-specific weaknesses that impact real-world applicability.

To address these challenges, we leverage multilingual transformer-based architectures and evaluate our model across diverse datasets to ensure robust generalization. We improve upon the current state-of-the-art by introducing a transliteration model using the m2m100 multilingual translation framework \cite{m2m100}. Our approach incorporates pretraining techniques, fine-tuning on multiple datasets, and Masked Language Modelling (MLM; introduced in \cite{bert}) to enhance transliteration quality. MLM pretraining improves the model’s ability to handle inconsistencies in Roman-Urdu spelling variations, enhancing transliteration robustness across datasets. To evaluate the effectiveness of our method, we assess performance on two datasets:
\begin{enumerate}
    \item Roman-Urdu-Parl (RUP) – A large-scale dataset of 6.3 million sentence pairs in Urdu and Roman-Urdu.
    \item Dakshina Dataset – A smaller, domain-specific dataset of 10,000 parallel sentences from Google.
\end{enumerate}

We establish new benchmarks by evaluating BLEU, Character BLEU, and CHRF scores for both Roman-Urdu → Urdu and Urdu → Roman-Urdu transliteration. Our results demonstrate substantial improvements, particularly in generalization across datasets, addressing a key limitation in previous research. 

Our key contributions are:
\begin{enumerate}
    \item Developing a Transformer-Based Transliteration Model: Fine-tuning m2m100 for Roman-Urdu transliteration, outperforming prior RNN-based approaches.
    \item Improving Domain Adaptation \& Generalization: Evaluating model performance on diverse datasets to ensure cross-domain robustness.
    \item Exploring MLM Pretraining: Applying masked language modeling (MLM) to improve model performance on informal Roman-Urdu word variations.
    
    
    \item Open-Sourcing Resources for Future Research: Providing trained models, evaluation scripts, and dataset processing pipelines to support future work in this domain \footnote{Models and datasets on \url{https://huggingface.co/Mavkif/}} 
    
\end{enumerate}

\section{Related Work}
Several studies explore techniques to improve translation and transliteration for languages with limited annotated corpora. A key driver of progress in data-driven NLP research has been the availability of large-scale datasets, as seen in machine translation with resources like OPUS \cite{opus}, WMT \cite{wmt}, and the United Nations Parallel Corpus \cite{undata}. However, most of these parallel corpora primarily support resource-rich languages, leaving low-resource languages underrepresented. Efforts to bridge this gap include parallel corpora for English-Hindi \cite{iit}, Nepali-English and Sinhala-English \cite{nepali}, and English with multiple Indian languages such as Bengali, Malayalam, Tamil and Telugu \cite{sixindian}. 

One significant approach to addressing data scarcity has been multilingual transfer learning. \cite{XLMR} introduced the XLM-R model, a multilingual language representation system trained on 100 languages, demonstrating that high-resource languages can improve performance for low-resource counterparts. Our work extends this approach by applying XLM-R to the task of transliteration between Roman Urdu and Urdu, highlighting the potential of multilingual models to bridge data limitations in low-resource settings.

Transliteration for languages with Arabic-derived scripts—such as Urdu, Persian, and Arabic has received limited attention compared to standard translation tasks. Notable work on Arabizi, the Romanized form of Arabic, highlights the relevance of transliteration research in other low-resource, informally written languages \cite{hajbi2024moroccan}. However, Roman-Urdu transliteration remains underexplored. The Roman-Urdu-Parl corpus \cite{Roman-urdu-parl}, a large-scale parallel dataset of 6.37 million sentence pairs, has served as a benchmark in this domain. Most existing systems based on this dataset rely on recurrent neural networks (RNNs) or statistical models. These methods face limitations in generalization and domain robustness, particularly due to the high variability and informality of Roman-Urdu text.

One challenge in Roman-Urdu-Parl is its test set construction. The dataset was augmented with Roman-Urdu variants for 5,000 frequent words (adding over 4 million pairs), but the test set was randomly sampled afterwards. As a result, models may encounter near-duplicate examples in training and testing, potentially inflating BLEU scores. Although randomization was assumed to mitigate this issue, separating test samples before augmentation would have ensured cleaner evaluation. Additionally, while a BLEU score \cite{bleu} of 84 was reported, character-level metrics such as Char BLEU \cite{charbleu} and CHRF \cite{chrf} are more suitable for transliteration due to the nature of the task.

Additionally, the Dakshina dataset \cite{dakshina} has been utilized for training and evaluation of transliteration systems for South Asian languages, including Urdu. Transformer-based architectures have shown improved character-level accuracy on this dataset compared to traditional models \cite{clsril}. These findings underscore the benefit of using more recent architectures and character-focused metrics in transliteration evaluation.

Furthermore, studies on cross-lingual transfer learning have shown promising results for low-resource NLP tasks \cite{crosslingual_massive}. By leveraging multilingual models pre-trained on high-resource languages, researchers have demonstrated improved performance in text generation and translation tasks for underrepresented languages. 

Another transliteration challenge arises in code-mixed settings common in Urdu and Roman Urdu, where English words appear within otherwise monolingual text. Simple character-level mapping is often insufficient for handling such cases. Prior studies \cite{code_switch} emphasize the importance of contextual modeling to correctly interpret whether a word should be transliterated or left intact. Context-aware multilingual models, especially those with sentence-level embeddings, have the potential to address these nuances effectively.

\section{Background on Romanization \& Roman Urdu}
Many multilingual communities, particularly in South Asia and Africa, use the Latin script (English alphabet) to write their native languages—a practice known as romanization. This phenomenon emerged due to early computing and mobile technology limitations, where English-based keyboards and software lacked support for native scripts. As a result, speakers of languages such as Urdu, Hindi, Bengali, and Arabic adapted by using Roman characters to represent their native words. This trend was further reinforced by the rise of the internet, where English remains the dominant digital language, making Romanized text a convenient alternative for communication.

\begin{figure}[t]
  \includegraphics[width=\columnwidth]{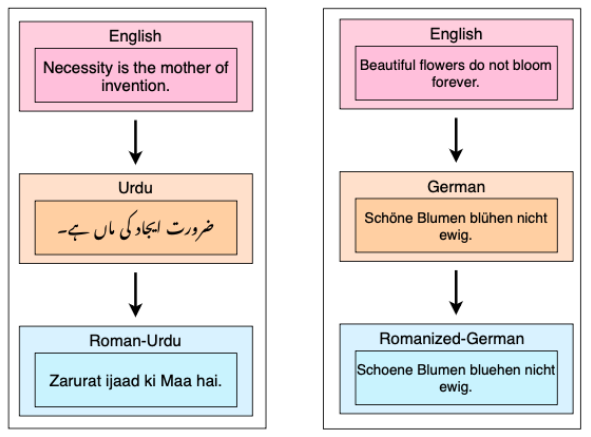}
  \caption{Examples of Romanization in English → Urdu and German → English. In both cases, the romanized sentences closely resemble the phonetic pronunciation in the native language.}
  \label{fig:romanization}
\end{figure}

Among these, Roman Urdu, a Romanized form of Urdu, has gained widespread adoption in social media, e-commerce, online news, and digital communication platforms. Unlike standardized Latin-based writing systems, however, Roman Urdu lacks a formal spelling convention, leading to highly inconsistent and phonetic spellings. The same word may be written in multiple ways, depending on how a speaker perceives and pronounces it. For example, as also shown in figure \ref{fig:romanization} the Urdu word for "what" can appear as "kia," "kya," or "keeya" in Roman Urdu, depending on the writer's preference.

This lack of standardization poses significant challenges for NLP and transliteration models, as they must handle spelling variations, informal grammar, and code-switching with English. Consequently, building an accurate and robust Roman-Urdu to Urdu transliteration system requires not only data-driven learning approaches but also an understanding of phonetic spelling inconsistencies and context-aware modeling.

\section{Methodology}
This section outlines the approach used to develop a Roman-Urdu and Urdu transliteration model based on the m2m100 multilingual translation model. Initially, we fine-tuned m2m100 directly on parallel transliteration data from the Roman-Urdu-Parl (RUP) dataset. While this yielded reasonable results, we observed that the model struggled with generalization when evaluated on a dataset it was not trained on, specifically the Dakshina dataset, which contains Roman-Urdu text from a different domain. When fine-tuned solely on RUP, the model performed well on RUP but struggled on other transliteration datasets such as the Google Dakshina dataset due to the variations in writing style in Roman-Urdu.

To mitigate this issue, we experimented with further fine-tuning on Dakshina, but this led to a considerable drop in performance on RUP, highlighting a trade-off between domain adaptation and dataset retention. This suggested that the model was overfitting to the dataset it was last trained on, rather than developing a generalizable transliteration capability. We also experimented with combining both datasets for training, but this approach was ineffective due to the large size difference between them. As discussed in Section 4.3 and shown in Section 5, we addressed this issue using MLM pretraining.

Overall, our methodology consists of two key phases: 
\begin{enumerate}
    \item Masked Language Modeling (MLM) pretraining, aimed at improving the model’s generalization for script-based variations
    \item fine-tuning for transliteration, where the model learns direct mappings between Roman-Urdu and Urdu
\end{enumerate}

Below, we describe the dataset split correction, model modifications, training phases, and hyperparameter settings in detail.

\subsection{Correcting the Roman-Urdu-Parl Dataset Split}
As discussed in Section 2, prior work on the Roman-Urdu-Parl dataset had issues with the evaluation methodology. We will address the issue and how we resolved it in this section. A closer examination of the dataset revealed that its train-test split could introduce data leakage, potentially inflating evaluation metrics. Specifically, the dataset was expanded by adding multiple Roman-Urdu variations for the 5,000 most frequent words. Since test data was randomly selected after these augmentations, models trained on this dataset may have seen structurally similar sentences in both training and test sets, differing only in minor spelling variations. This likely led to overestimated BLEU scores, as the model was not truly generalizing but rather memorizing transliteration variants seen during training. To address this, we implemented a structured dataset split to ensure that no Urdu sentence (or its Roman-Urdu variations) appeared in more than one subset (training, validation, or test).

\subsubsection{Dataset Splitting Strategy}
To assess transliteration quality on sentences with only one Roman-Urdu variant, we extracted 1,500 Urdu sentences that appear only once in the dataset for the validation set, along with their corresponding Roman-Urdu text. Another 1,500 unique Urdu sentences were selected for the test set, ensuring no overlap with validation. 

Similarly, we also included variation-rich transliterations in all the sets. we selected Urdu sentences that had between 2 and 10 transliterations. The validation set contains 3,000 Urdu sentences with multiple transliteration variations, including all their variations. Another 3,000 Urdu sentences were assigned to the test set, ensuring that all variations for these sentences appear only in the test set and not in validation. This split ensures that the model is evaluated on both simple and complex transliteration cases.

The training set was formed by removing all validation and test sentences and their variations from the dataset. This guaranteed no overlap between training and evaluation sets, preventing data leakage while maintaining a diverse training corpus that covers a wide range of transliteration styles.

To facilitate faster validation and testing, we created smaller validation and test subsets, each containing 4,500 sentences. A subset of 1,500 unique Urdu sentences was selected from the full validation and test sets, while for replicated sentences, a randomly selected transliteration was chosen from each multi-variant Urdu sentence, reducing redundancy. These smaller subsets enable efficient model evaluation while maintaining dataset integrity and diversity. Validation and testing of the models were done

\subsubsection{Ensuring Data Integrity and Generalization}
To confirm the reliability of our dataset split, we implemented multiple data integrity checks:
\begin{enumerate}
    \item No Overlap Between Sets: No Urdu sentence (or its variations) appears in more than one subset, ensuring a fair evaluation.
    \item Variation Inclusion: The full test and validation sets include all variations of selected Urdu sentences, ensuring a robust assessment.
    \item Balanced Evaluation: The dataset contains both unique and multi-variant transliterations, providing a comprehensive evaluation.
    \item Training Set Integrity Check: No training sentence was found to be a duplicate or partial repetition of test or validation sentences.    
\end{enumerate}

By restructuring the dataset with this non-overlapping split, we ensure that our model evaluations reflect true generalization rather than memorization, setting a more rigorous benchmark for transliteration tasks.

\subsection{Model Architecture and Tokenization}
We use m2m100, a transformer-based sequence-to-sequence model, as the backbone for transliteration. Since m2m100 does not include Roman-Urdu as a distinct language, we extended its tokenizer by introducing language-specific tokens for Roman-Urdu (\_\_roman-ur\_\_) and Urdu (\_\_ur\_\_). This modification enabled explicit language conditioning during training and inference.
All the text was tokenized using the modified tokenizer, and sequences were truncated or padded to a maximum length of 128 tokens. This ensures compatibility with the model’s architecture while preserving transliteration accuracy.

\subsection{Masked Language Modeling (MLM) Pretraining}
Due to the domain adaption issue that we discussed earlier in this section, We hypothesized that explicit pretraining on Roman-Urdu and Urdu as separate monolingual corpora would help the model internalize subword structures and script-based variations before learning direct transliteration mappings. This would allow it to develop a more generalized understanding of character and subword transformations, making it less dependent on the specific dataset used for fine-tuning. To test this hypothesis, we introduced an MLM pretraining phase, where the model was trained to reconstruct masked tokens in monolingual Roman-Urdu and Urdu text. By exposing the model to a diverse range of Roman-Urdu spellings and Urdu orthographic patterns, we aimed to create a stronger foundational representation, improving cross-dataset generalization during the final transliteration fine-tuning stage. 
For training data, we used both the Roman-Urdu-Parl and Google Dakshina datasets, as they are among the largest available corpora containing Roman-Urdu text. Since MLM pretraining only uses raw text rather than parallel text, incorporating the same datasets for both MLM and transliteration fine-tuning did not pose an issue. This approach ensured that the model encountered a rich variety of linguistic patterns during pretraining while maintaining consistency with the fine-tuning data.

\subsubsection{Pretraining Data and Training Setup}
Two MLM pretraining configurations were used:
\begin{enumerate}
    \item Roman-Urdu Only MLM: The model was trained on Roman-Urdu monolingual text, where 15\% of tokens were randomly masked and reconstructed by the model.
    \item Roman-Urdu + Urdu MLM: The training data consisted of both Roman-Urdu and Urdu text, effectively doubling the corpus size and allowing the model to learn script-level relationships.
\end{enumerate}
During MLM pretraining, the initial encoder and decoder layers were frozen to retain the multilingual translation capabilities of m2m100. Specifically, embedding layers, positional encoding layers, and the first two transformer layers in both the encoder and decoder were kept frozen. This ensured that only the deeper layers adapted to Roman-Urdu and Urdu representations. For the hyper parameters for this task, we used a maximum sequence length of 128 tokens, ensuring consistency in input representation. The model was trained for 4 epochs, using a batch size of 128 for both training and evaluation. To optimize memory usage, gradient accumulation was set to 4 steps, effectively increasing the batch size without exceeding hardware limitations.

\subsection{Fine-Tuning for Transliteration}
Following MLM pretraining, the model was fine-tuned on parallel Roman-Urdu and Urdu data for direct sequence-to-sequence transliteration. The training corpus consisted of Roman-Urdu-Parl and Google's Dakshina Dataset. 
Unlike MLM pretraining, no layers were frozen during fine-tuning, allowing the model to fully adjust its parameters to transliteration-specific patterns. The language-specific tokens (\_\_roman-ur\_\_, \_\_ur\_\_) were used throughout to condition the model on the source and target scripts.
Initially, the model was trained solely on Roman-Urdu-Parl for 15 epochs, but performance gains plateaued after 5 epochs, indicating diminishing returns beyond this point.
To improve cross-domain generalization and make the model suitable for different datasets, the model was further fine-tuned on the Google Dakshina dataset, which consists of 10,000 Roman-Urdu and Urdu sentence pairs from a different linguistic domain. However, we observed that extensive fine-tuning on Dakshina degraded performance on RUP, suggesting a trade-off between domain adaptation and retention of previously learned transliteration patterns. To mitigate this, we fine-tuned the 5th epoch checkpoint from RUP training on Dakshina for up to 5 additional epochs. Given the observed BLEU score degradation on RUP, we compared performance at 2 and 5 epochs of Dakshina fine-tuning, balancing improvements in domain adaptation with retention of previously learned transliteration mappings.

\subsubsection{Training Configuration}
As discussed, Fine-tuning was conducted in two phases:
\begin{enumerate}
    \item Initial training on RUP: 15 epochs, with results plateauing after 5 epochs.
    \item Further fine-tuning on Dakshina: Using the 5th epoch checkpoint from RUP, fine-tuned for 2 to 5 additional epochs, evaluating the trade-off between domain adaptation and performance retention.
\end{enumerate}

In both training phases, batch size of 64 and gradient accumulation over 4 steps were used. A learning rate of 1e-5 with a 10\% warmup ratio was applied, ensuring a stable adaptation process. To prevent overfitting, a weight decay of 0.02 was used. For Roman-Urdu-Parl dataset testing, we used the test set of 4,500 samples. The larger dataset is still maintained for future references. 

\subsection{Evaluation Metrics Selection}
Evaluating transliteration models requires metrics that accurately capture character-level accuracy, as transliteration involves precise phonetic and orthographic mappings. Previous work on the Roman-Urdu-Parl (RUP) dataset primarily used n-gram BLEU scores, which are widely used for machine translation but may not fully reflect transliteration quality. Since transliteration operates at a subword or character level, n-gram BLEU can fail to capture finer-grained spelling variations, making it less suited for evaluating transliteration tasks where minor character-level differences significantly impact correctness. We also include GPT-4o Mini as a zero-shot comparison to evaluate how general-purpose LLMs perform on this specialized transliteration task. For this we set the temperature and top\_p parameters to 1. Rest of the parameters were set to default values. 

To ensure a more comprehensive evaluation, we assessed our models using three metrics:
\begin{enumerate}
    \item General BLEU Score (N-Gram BLEU) – Used for comparison with prior work, as this was the metric reported in the RUP paper.
    \item Character-Level BLEU (Char BLEU) – A finer-grained metric that evaluates character sequences rather than word-based n-grams, making it more appropriate for transliteration tasks.
    \item Character F-score (CHRF) – A recall-oriented metric that captures partial character-level overlaps, which is useful for handling spelling variations in transliteration.
\end{enumerate}

While we evaluated all models using all three metrics, we report only character-level BLEU (Char BLEU) in the results section. This is because Char BLEU provides the most reliable indication of transliteration accuracy, as it directly evaluates character sequences rather than word-based n-grams. Given the high spelling variation in Roman-Urdu, the general BLEU metric can over-penalize transliterations that differ slightly in spelling but remain phonetically correct, making Char BLEU a more meaningful measure. CHRF, while useful for handling partial character overlaps, showed trends similar to Char BLEU, making the char BLEU the primary focus of our analysis.

By prioritizing Char BLEU in our evaluation, we ensure that the reported results align with the nature of transliteration tasks, where fine-grained character accuracy is more important than broader phrase-level overlap.

\section{Results and Discussion}
In this section, we present the results of our Roman-Urdu and Urdu transliteration experiments, comparing models trained with and without MLM pretraining. Character-level BLEU (Char-BLEU) is used as our primary evaluation metric in the main tables.
While additional BLEU and CHRF scores are reported in the appendix \ref{sec:appendix}

\begin{table*}
  \centering
    \begin{tabular}{|l|c|}
        \hline
        \textbf{Method} & \textbf{BLEU Score} \\
        \hline
        Roman-Urdu-Parl (RNN+LSTM) & 84.67 \\
        GPT-4o Mini (zero-shot) & 80.966 \\
        Our Work & \textbf{94.586} \\
        \hline
    \end{tabular}
  \caption{\label{tab:bleu_comparison}
    4-gram BLEU score comparison for Roman-Urdu → Urdu transliteration, as reported in the RUP paper, GPT-4o Mini, and our work.
  }
\end{table*}

\begin{table*}[ht]
    \centering
    \begin{tabular}{|l|l|l|c|c|}
    \hline
    \textbf{Model Variant} & \textbf{Trained on} & \textbf{Tested on} & \textbf{Without MLM} & \textbf{With MLM} \\
    \hline
    Fine-tuned m2m100 & RUP & RUP & 97.29 & 97.44 \\
    Fine-tuned m2m100 & RUP + Dakshina & RUP & 93.72 & 97.06 \\
    GPT-4o Mini & – & RUP & 80.96 & 80.96 \\
    Fine-tuned m2m100 & RUP & Dakshina & 81.96 & 83.11 \\
    Fine-tuned m2m100 & RUP + Dakshina & Dakshina & 91.21 & 91.34 \\
    GPT-4o Mini & – & Dakshina & 92.02 & 92.02 \\
    \hline
    \end{tabular}
    \caption{Char-BLEU scores for Roman-Urdu $\rightarrow$ Urdu transliteration with and without MLM pretraining.}
    \label{tab:rur_ur_charbleu}
\end{table*}

\begin{table*}[ht]
    \centering
    \begin{tabular}{|l|l|l|c|c|}
    \hline
    \textbf{Model Variant} & \textbf{Trained on} & \textbf{Tested on} & \textbf{Without MLM} & \textbf{With MLM} \\
    \hline
    Fine-tuned m2m100 & RUP & RUP & 96.36 & 96.37 \\
    Fine-tuned m2m100 & RUP + Dakshina & RUP & 85.05 & 86.39 \\
    GPT-4o Mini & – & RUP & 67.90 & 67.90 \\
    Fine-tuned m2m100 & RUP & Dakshina & 68.68 & 71.27 \\
    Fine-tuned m2m100 & RUP + Dakshina & Dakshina & 76.89 & 78.45 \\
    GPT-4o Mini & – & Dakshina & 75.06 & 75.06 \\
    \hline
    \end{tabular}
    \caption{Char-BLEU scores for Urdu $\rightarrow$ Roman-Urdu transliteration with and without MLM pretraining.}
    \label{tab:ur_rur_charbleu}
\end{table*}

We begin by comparing our Roman-Urdu → Urdu transliteration performance with baseline results reported in the Roman-Urdu-Parl paper. As shown in Table ~\ref{tab:bleu_comparison}, the previous RNN+LSTM model achieved a BLEU score of 84.67, while GPT-4o Mini, evaluated in a zero-shot setting, scored 80.96. In contrast, our fine-tuned m2m100 model trained solely on Roman-Urdu-Parl data achieves a BLEU score of 94.58, setting a new benchmark for this task and highlighting the effectiveness of transformer-based models over traditional sequence-to-sequence architectures. We selected Bleu score for this comparison because the baseline results reported used this metric, and the model was not publically available.

Tables ~\ref{tab:rur_ur_charbleu} and ~\ref{tab:ur_rur_charbleu} present detailed Char-BLEU results for Roman-Urdu → Urdu and Urdu → Roman-Urdu directions, respectively. Without MLM pretraining, models trained on RUP perform well on in-domain test data but struggle to generalize to the Dakshina dataset. Fine-tuning on Dakshina improves performance on out-of-domain data but leads to some degradation on RUP, reflecting a trade-off between domain adaptation and retention of previously learned mappings.

We observe that MLM pretraining helps mitigate this trade-off. While the overall performance gain is modest, MLM-pretrained models demonstrate more balanced performance across datasets. In particular, fine-tuning on Dakshina after MLM pretraining results in improved generalization without significantly harming performance on RUP, suggesting that MLM helps stabilize learning by exposing the model to diverse subword and script-level patterns in advance.

To provide additional context, we also evaluated OpenAI's GPT-4o Mini in a zero-shot setup using prompt-based decoding. GPT-4o serves as a strong multilingual baseline, having been trained on diverse transliteration patterns across multiple languages. While it performs reasonably well on Dakshina, it lags behind our fine-tuned model on RUP. Notably, in the Urdu → Roman-Urdu direction, GPT-4o is more competitive, likely due to its broader exposure to Romanized scripts and informal user-generated text.

Overall, we find that Roman-Urdu → Urdu transliteration is consistently easier for models, likely due to the standardized nature of Urdu script. In contrast, Urdu → Roman-Urdu transliteration is inherently ambiguous due to the lack of orthographic standardization in Roman-Urdu, where multiple phonetic spellings may exist for the same word. While Char-BLEU offers a more forgiving measure than n-gram BLEU, it still penalizes plausible variants not present in the reference. Addressing this limitation—by incorporating multi-reference evaluation or designing more flexible scoring metrics—remains a promising direction for future work.

In summary, our results demonstrate that transformer-based models fine-tuned on domain-specific data outperform both traditional RNNs and large-scale general-purpose LLMs like GPT-4o. MLM pretraining contributes to better stability and cross-domain robustness, particularly in low-resource, non-standardized transliteration settings.

\section{Conclusion and Future Work}
This paper presents a fine-tuned transliteration model for Roman-Urdu and Urdu, based on the m2m100 multilingual translation model with Masked Language Modeling (MLM) pretraining. Our results show that MLM pretraining significantly enhances transliteration performance, improving cross-dataset generalization while maintaining strong performance on the primary dataset (RUP). Fine-tuning on Roman-Urdu-Parl (RUP) allows our model to surpass previous state-of-the-art (SOTA) results and GPT-4o Mini, setting a new benchmark for Roman-Urdu to Urdu transliteration. These findings highlight the effectiveness of transformer-based models over traditional sequence-to-sequence architectures and the importance of domain-specific training.

Future work includes exploring multiple reference transliterations to better handle Roman-Urdu spelling variability. Transliteration research for non-standardized Romanized scripts can also benefit from MLM pretraining, particularly when no dedicated training data is available. Applying this approach to other South Asian languages, such as Hindi or Bengali, could further validate its broader applicability. Another promising direction is semi-supervised learning, where the model leverages unlabeled Roman-Urdu text to reduce reliance on parallel corpora.

\section{Acknowledgement}
This research was conducted at the German Research Center for Artificial Intelligence (DFKI) and the University of Saarland. The authors gratefully acknowledge DFKI for providing the necessary hardware resources and a supportive research environment.  This work was also supported by the German Federal Ministry of Education and Research (BMBF) as part of the TRAILS project (01IW24005). The authors thank the BMBF for their commitment to advancing research in Artificial Intelligence.

\newpage

\bibliography{custom}

\newpage

\appendix
\section{Appendix}
\label{sec:appendix}

\subsection*{Training Setup and Table Column Definitions}

\subsubsection*{1. Overview}

\textbf{Objective:} Improve the general performance of a transliteration model between Roman-Urdu (roman-ur) and Urdu (ur) across different datasets and domains.

\textbf{Model Used:} Fine-tuned the m2m100 multilingual translation model.

\textbf{Datasets Used:}
\begin{itemize}
  \item \textbf{Primary Dataset (RUP):} 6.3 million sentence pairs of Roman-Urdu and Urdu text collected from various informal online sources.
  \item \textbf{Secondary Dataset (Dakshina):} Google's Dakshina dataset with 10,000 Roman-Urdu/Urdu sentence pairs sourced from Wikipedia. This was used to improve domain generalization after observing poor performance (~50 BLEU) on it when using only RUP-trained models.
\end{itemize}

\subsubsection*{2. Training and Fine-Tuning Procedure}

\begin{itemize}
  \item Initial training was conducted on the RUP dataset for both transliteration directions (roman-ur → ur and ur → roman-ur), achieving high BLEU scores (~90) on the RUP test set.
  \item The model exhibited weaknesses on out-of-domain data (e.g., Dakshina), particularly with proper names and formal text.
  \item Further fine-tuning was done on the Dakshina dataset after initial RUP training. Results are shown after the 2nd and 5th epochs of training on Dakshina. 

  \item Due to the significant size imbalance between RUP and Dakshina, joint training would cause the model to underfit the smaller Dakshina domain, leading to poor generalization; hence, we first trained on RUP for general transliteration capability and then fine-tuned on Dakshina to specialize for that domain.
  \item Additionally, we experimented with masked language modeling (text infilling) as a form of pretraining using the raw text from both datasets. Two variants were used:
  \begin{enumerate}
    \item Pretrained only on Roman-Urdu text.
    \item Pretrained on both Urdu and Roman-Urdu text (double the size).
  \end{enumerate}
  \item Most transformer layers were frozen during infilling pretraining.
\end{itemize}

\subsubsection*{3. Table Column Definitions}

The results table includes performance scores across various model configurations and evaluation sets. Each row represents a specific model configuration.

\textbf{Columns:}
\begin{itemize}
  \item \textbf{Configuration:} Indicates whether the model was pretrained using Roman-Urdu infilling, and which variant.
  \item \textbf{Model:} Direction of transliteration (roman-ur → ur or ur → roman-ur).
  \item \textbf{Epoch:} 
  \begin{itemize}
    \item "Without Further Finetuning": The model was only trained on the RUP dataset.
    \item "2nd", "5th": Epoch number of additional fine-tuning on the Dakshina dataset.
  \end{itemize}
  \item \textbf{GPT4o-mini BLEU, Char BLEU, CHRF (RUP/Dakshina):} Scores from OpenAI GPT-4o mini API on the same evaluation sets, included for reference. These values are repeated across rows since the API is not fine-tuned per configuration.
  \item \textbf{Our Model Scores:} BLEU, CharBLEU, and CHRF scores of our models on both the RUP and Dakshina test sets.
\end{itemize}

\textbf{Test Set Sizes:}
\begin{itemize}
  \item \textbf{RUP Test Set:} 4,500 examples
  \item \textbf{Dakshina Test Set:} 500 examples
\end{itemize}

\textbf{Note:} While the main paper focuses on BLEU and CharBLEU results, this table provides detailed scores for all three metrics 4-gram BLEU, Character-level BLEU, and CHRF for all configurations, ensuring full transparency.

\begin{sidewaystable}
\centering
\resizebox{\textheight}{!}{%
\begin{tabular}{lllrrrrrrrrrrrrr}
\toprule
Configuration & Model & Epoch & \multicolumn{3}{c}{GPT4o-mini BLEU} & \multicolumn{3}{c}{GPT4o-mini CHRF} & \multicolumn{6}{c}{Our Model Scores} \\
\cmidrule(lr){4-6} \cmidrule(lr){7-9} \cmidrule(lr){10-15}
& & & RUP & Dakshina & RUP Char & Dakshina Char & RUP & Dakshina & RUP BLEU & Dakshina BLEU & RUP Char BLEU & Dakshina Char BLEU & RUP CHRF & Dakshina CHRF \\
\midrule
Without Roman-Urdu Infil & Roman-Urdu → Urdu & Without Finetuning & 57.535 & 75.254 & 80.966 & 92.029 & 76.50 & 89.58 & 94.586 & 53.089 & 97.296 & 81.916 & 96.73 & 77.48 \\
Without Roman-Urdu Infil & Roman-Urdu → Urdu & 2nd & 57.535 & 75.254 & 80.966 & 92.029 & 76.50 & 89.58 & 91.721 & 70.341 & 96.435 & 89.102 & 95.67 & 86.08 \\
Without Roman-Urdu Infil & Roman-Urdu → Urdu & 5th & 57.535 & 75.254 & 80.966 & 92.029 & 76.50 & 89.58 & 83.935 & 74.825 & 93.732 & 91.212 & 91.93 & 88.66 \\
Without Roman-Urdu Infil & Urdu → Roman-Urdu & Without Finetuning & 27.878 & 31.254 & 67.908 & 75.061 & 62.90 & 67.96 & 88.781 & 24.867 & 96.365 & 68.681 & 95.31 & 61.84 \\
Without Roman-Urdu Infil & Urdu → Roman-Urdu & 2nd & 27.878 & 31.254 & 67.908 & 75.061 & 62.90 & 67.96 & 75.512 & 33.937 & 89.926 & 76.110 & 88.25 & 69.20 \\
Without Roman-Urdu Infil & Urdu → Roman-Urdu & 5th & 27.878 & 31.254 & 67.908 & 75.061 & 62.90 & 67.96 & 63.066 & 36.849 & 85.050 & 76.897 & 82.48 & 70.33 \\

With Roman-Urdu Infil & Roman-Urdu → Urdu & Without Further Finetuning & 57.535 & 75.254 & 80.966 & 92.029 & 76.50 & 89.58 & 94.801 & 53.060 & 97.306 & 82.255 & 96.75 & 77.76 \\
With Roman-Urdu Infil & Roman-Urdu → Urdu & 2nd & 57.535 & 75.254 & 80.966 & 92.029 & 76.50 & 89.58 & 93.307 & 70.979 & 96.621 & 89.212 & 95.92 & 86.19 \\
With Roman-Urdu Infil & Roman-Urdu → Urdu & 5th & 57.535 & 75.254 & 80.966 & 92.029 & 76.50 & 89.58 & 90.747 & 74.280 & 95.835 & 90.826 & 94.87 & 88.21 \\
With Roman-Urdu Infil & Urdu → Roman-Urdu & Without Further Finetuning & 27.878 & 31.254 & 67.908 & 75.061 & 62.90 & 67.96 & 88.848 & 25.759 & 96.425 & 69.310 & 95.41 & 62.29 \\
With Roman-Urdu Infil & Urdu → Roman-Urdu & 2nd & 27.878 & 31.254 & 67.908 & 75.061 & 62.90 & 67.96 & 73.045 & 33.905 & 88.520 & 75.835 & 86.88 & 68.95 \\
With Roman-Urdu Infil & Urdu → Roman-Urdu & 5th & 27.878 & 31.254 & 67.908 & 75.061 & 62.90 & 67.96 & 63.574 & 36.987 & 85.088 & 78.149 & 82.59 & 71.55 \\

With Urdu + Roman-Urdu Infil & Roman-Urdu → Urdu & Without Finetuning & 57.535 & 75.254 & 80.966 & 92.029 & 76.50 & 89.58 & 94.862 & 52.978 & 97.442 & 83.119 & 96.89 & 78.75 \\
With Urdu + Roman-Urdu Infil & Roman-Urdu → Urdu & 2nd & 57.535 & 75.254 & 80.966 & 92.029 & 76.50 & 89.58 & 94.630 & 72.514 & 97.260 & 89.965 & 96.71 & 87.12 \\
With Urdu + Roman-Urdu Infil & Roman-Urdu → Urdu & 5th & 57.535 & 75.254 & 80.966 & 92.029 & 76.50 & 89.58 & 94.076 & 75.598 & 97.076 & 91.345 & 96.49 & 88.84 \\
With Urdu + Roman-Urdu Infil & Urdu → Roman-Urdu & Without Finetuning & 27.878 & 31.254 & 67.908 & 75.061 & 62.90 & 67.96 & 88.733 & 28.054 & 96.376 & 71.274 & 95.33 & 64.24 \\
With Urdu + Roman-Urdu Infil & Urdu → Roman-Urdu & 2nd & 27.878 & 31.254 & 67.908 & 75.061 & 62.90 & 67.96 & 78.370 & 34.681 & 90.976 & 76.222 & 89.56 & 69.26 \\
With Urdu + Roman-Urdu Infil & Urdu → Roman-Urdu & 5th & 27.878 & 31.254 & 67.908 & 75.061 & 62.90 & 67.96 & 67.226 & 37.969 & 86.391 & 78.452 & 84.07 & 71.83 \\

\bottomrule
\end{tabular}%
}
\caption{Evaluation results for Roman-Urdu to Urdu transliteration with and without MLM and further fine-tuning.}
\end{sidewaystable}

\end{document}